\documentclass[journal]{IEEEtran}
\usepackage{cite}

\ifCLASSINFOpdf
  \usepackage[pdftex]{graphicx}
\else
\fi
\usepackage{tikz}
\usepackage{pgfplots}
\usepackage{mathtools}

\usetikzlibrary{calc,trees,positioning,arrows,chains,shapes.geometric,%
    decorations.pathreplacing,decorations.pathmorphing,shapes,%
    matrix,shapes.symbols,snakes}

\tikzset{%
  block/.style  = {draw, thick, rectangle, minimum height = 3em, minimum width = 3em},
  neuron/.style = {draw, circle, minimum height=2em},
  sum/.style    = {draw, circle}, 
  layer/.style  = {rectangle, draw=black, minimum height=1.7em, minimum width=15mm, text centered},
}
\tikzset{%
  tnode/.style   = {draw, circle, minimum height=0.6em,inner sep=0},
  rnode/.style   = {draw, circle, minimum height=0.3em,
                    fill=black, inner sep=0},
  rollout/.style   = {>=stealth,snake=coil,segment aspect=0},
}

\usepackage{amsmath,amssymb}
\newcommand{\argmax}{\mathop{\mathrm{argmax}}}
\DeclareMathOperator{\E}{\mathbb{E}}
\DeclareMathOperator{\UCB}{UCB}
\DeclareMathOperator{\Beta}{Beta}

\usepackage{algorithm, algorithmic}

\usepackage{listings}
\usepackage{xcolor}

\definecolor{codegreen}{rgb}{0,0.6,0}
\definecolor{codegray}{rgb}{0.5,0.5,0.5}
\definecolor{codepurple}{rgb}{0.58,0,0.82}
\definecolor{backcolour}{rgb}{0.95,0.95,0.92}

\lstdefinestyle{mystyle}{
    backgroundcolor=\color{backcolour},   
    commentstyle=\color{codegreen},
    keywordstyle=\color{magenta},
    numberstyle=\tiny\color{codegray},
    stringstyle=\color{codepurple},
    basicstyle=\ttfamily\footnotesize,
    breakatwhitespace=false,         
    breaklines=true,
    keepspaces=true,                 
    numbers=none,
    numbersep=5pt,                  
    showspaces=false,                
    showstringspaces=false,
    showtabs=false,                  
    tabsize=2
}
\makeatletter
\def\lst@makecaption{%
  \def\@captype{table}%
  \@makecaption
}
\makeatother

\definecolor{diffcolor}{rgb}{0.0, 0.0, 0.0}

\usepackage{url}

\usepackage{booktabs}
\usepackage{multirow}

\begin{document}
%
\title{Learning to Play Imperfect-Information Games\\ by Imitating an Oracle Planner}
%
%
%

\author{Rinu Boney, Alexander Ilin, Juho Kannala and Jarno Seppänen%
\thanks{R. Boney*, A. Ilin and J. Kannala are with Department of Computer Science, Aalto University, Espoo, Finland; J. Seppänen is with Supercell, Helsinki, Finland.}
\thanks{* work done as an Intern at Supercell.}
}

\maketitle

\begin{abstract}
We consider learning to play multiplayer imperfect-information games with simultaneous moves and large state-action spaces. Previous attempts to tackle such challenging games have largely focused on model-free learning methods, often requiring hundreds of years of experience to produce competitive agents. Our approach is based on model-based planning. We tackle the problem of partial observability by first building an (oracle) planner that has access to the full state of the environment and then distilling the knowledge of the oracle to a (follower) agent which is trained to play the imperfect-information game by imitating the oracle's choices. We experimentally show that planning with naive Monte Carlo tree search does not perform very well in large combinatorial action spaces. We therefore propose planning with a fixed-depth tree search and decoupled Thompson sampling for action selection. We show that the planner is able to discover efficient playing strategies in the games of Clash Royale and Pommerman and the follower policy successfully learns to implement them by training on a few hundred battles.
\end{abstract}

\begin{IEEEkeywords}
Imperfect-information games, Monte Carlo tree search, Clash Royale, Pommerman.
\end{IEEEkeywords}

%
\IEEEpeerreviewmaketitle

\section{Introduction}
%
%
%
%

 

\IEEEPARstart{T}{he} goal of the field of reinforcement learning (RL) is to develop learning algorithms that can effectively deal with the complexities of the real world. Games are a structured form of interactions between one or more players in an environment, making them ideal for the study of reinforcement learning. Much of research in artificial intelligence has focused on games which emulate different challenges of the real world. In Go \cite{silver2017mastering}, the agent 
has to discover complex strategies in a large search space.
In card games like Poker \cite{moravvcik2017deepstack, brown2018superhuman, brown2019superhuman}, the agent has to deal with the imperfect-information, such as the unknown cards of the opponent. In StarCraft~II~\cite{vinyals2019grandmaster} and Dota~2~\cite{berner2019dota}, the agent has to compete with other agents who take simultaneous actions from a large action space.

\color{diffcolor}

In this work, we consider the problem of learning to play games with a novel set of challenges: imperfect-information multi-agent games with simultaneous moves and large state-action spaces. 
We consider two such games as learning environments: Clash Royale (a popular multiplayer real-time strategy game) and Pommerman \cite{DBLP:journals/corr/abs-1809-07124}.
Clash Royale is a unique game combining elements of different genres such as MOBA (multiplayer online battle arena), collective-card games, and tower defense games. The complexity in learning to play Clash Royale comes from the presence of cyclic strategies, partial observability, and exploration in large dynamic action spaces (more details in Section~\ref{sec:cr}).
Pommerman is a popular multi-agent RL benchmark which is difficult due to the need for opponent modelling and therefore a large branching factor as decisions are made in the combinatorial action space.

\begin{figure}[t]
\centering
\includegraphics[width=\linewidth, trim=0 0 70 0, clip]{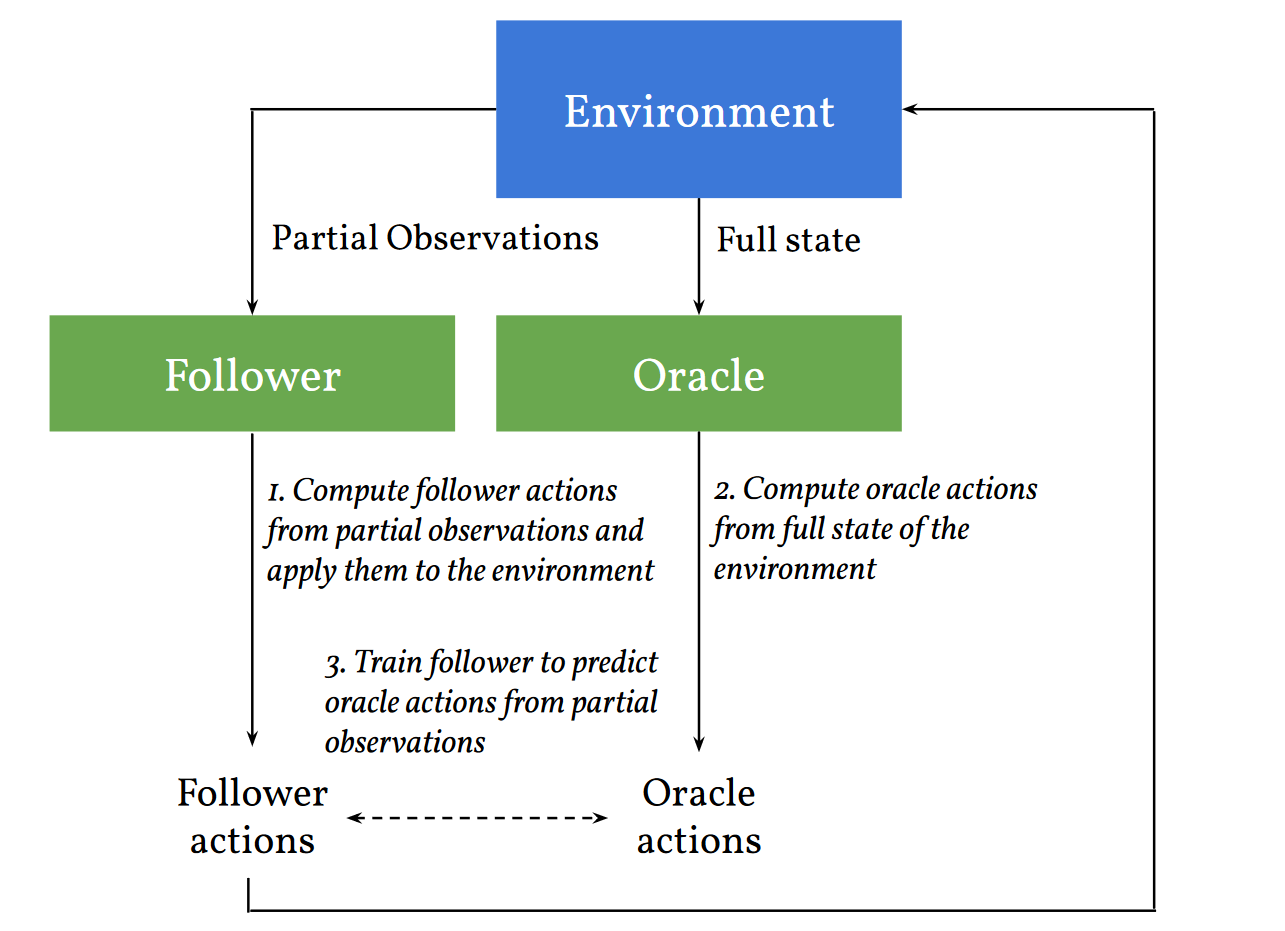}
\caption{\color{diffcolor} Proposed approach to solving imperfect-information games.}
\label{fig:approach}
\end{figure}

In this paper, we introduce a new algorithm for efficient learning in large imperfect-information games\footnote{See \url{https://sites.google.com/view/l2p-clash-royale} for an explanation video of our method, in the context of Clash Royale.}, which does not require modifying the core game implementation. Our approach (illustrated in Fig.~\ref{fig:approach}) consists of two separate components: an oracle planner and a follower agent. The oracle planner has access to the full state of the environment and performs self-play tree search to compute effective (oracle) actions for each player. The oracle planner by itself can be used to implement a cheating AI for game implementations that do not support randomizing hidden information.
A follower agent that can play the imperfect-information game is obtained by training a neural network to predict the oracle actions from partial observations using supervised learning. 

Planning is non-trivial in imperfect-information games \cite{brown2017safe}. The classical solution is to use Monte Carlo tree search (MCTS) with \emph{determinization} of the hidden information during search to account for the lack of the fully observed state of the environment \cite{frank1998search, ginsberg2001gib, bjarnason2009lower, cowling2012information}. However, this approach cannot be directly used in practice for many games as
most existing simulators do not support the possibility of varying the hidden information. 

\color{black}

Simultaneous moves with large action spaces makes model-based planning exceptionally challenging.
Conventional MCTS can easily get stuck at creating new nodes corresponding to untried actions in a combinatorial action space.
In this paper, we propose to build an oracle planner based on fixed-depth tree search (FDTS) with use of decoupled Thompson sampling for action selection. Our experiments show that FDTS can discover efficient strategies via self-play in the two challenging games that we consider in the paper.


\emph{Contributions.} 1)~We introduce a new algorithm for efficient planning and learning in large imperfect-information games with implementations that do not support varying of hidden information. 2)~We demonstrate that naive Monte Carlo tree search can be problematic in large action spaces and introduce fixed-depth tree search to improve the quality of planning. 3)~We demonstrate the effectiveness of the algorithm in the novel setting of Clash Royale and the popular multi-agent RL benchmark of Pommerman.

\lstset{basicstyle=\scriptsize\ttfamily,breaklines=true}

\section{Imperfect-Information Games}

We formalize imperfect-information games as partially-observable stochastic games (POSG) \cite{hansen2004dynamic}.
In POSG, a game is played by a set of $N$ players and each game begins in an initial state $s_0$ sampled from an initial state distribution. 
In any state $s$, observation functions $O_i(s)$ yield observations $o_i = O_i(s)$ for each player $i$. After receiving observation $o_i$, each player $i$ chooses an action $a_i \in A_i(s)$, where $A_i(s)$ is the set of actions available to player $i$ in state $s$.
Once all players choose actions $a = (a_1, \ldots, a_N)$, the game transitions to a new state $s'$ as defined by a transition function $s' = f(s, a)$.
Thus, the joint action space is $A(s) = A_1(s) \times \ldots \times A_N(s)$. 
The end of a game is defined by a set of terminal states $Z$. Once the game reaches a terminal state $z \in Z$, all players receive a ternary reward of 1 (win), 0 (draw) or -1 (loss) as defined by a reward function $R_i(z)$. A player does not have access to the true initial state distribution or the transition function but can sample from them by playing games. We now introduce the two games studied in this paper.

\subsection{Clash Royale}
\label{sec:cr}

Clash Royale is a multiplayer real-time strategy game consisting of short battles lasting a few minutes. We focus on the two-player mode of Clash Royale. 
Before a battle, each player picks a \emph{deck} of eight different cards that is not revealed to the opponent.
The game has nearly 100 \emph{cards} that represents playable troops, buildings or spells that will be used in battles. 
As the game begins, each player is dealt a random subset of four different cards (\emph{hand}) from their deck.
Solving the whole game of Clash Royale involves solving the meta-game of choosing the right deck.
In this paper, we focus on a fixed beginner deck consisting of \emph{Knight, Giant, Archer, Arrows, Minions, Fireball, Musketeer,} and \emph{Baby Dragon}.

\begin{figure}[!t]
\centering
\includegraphics[height=50mm]{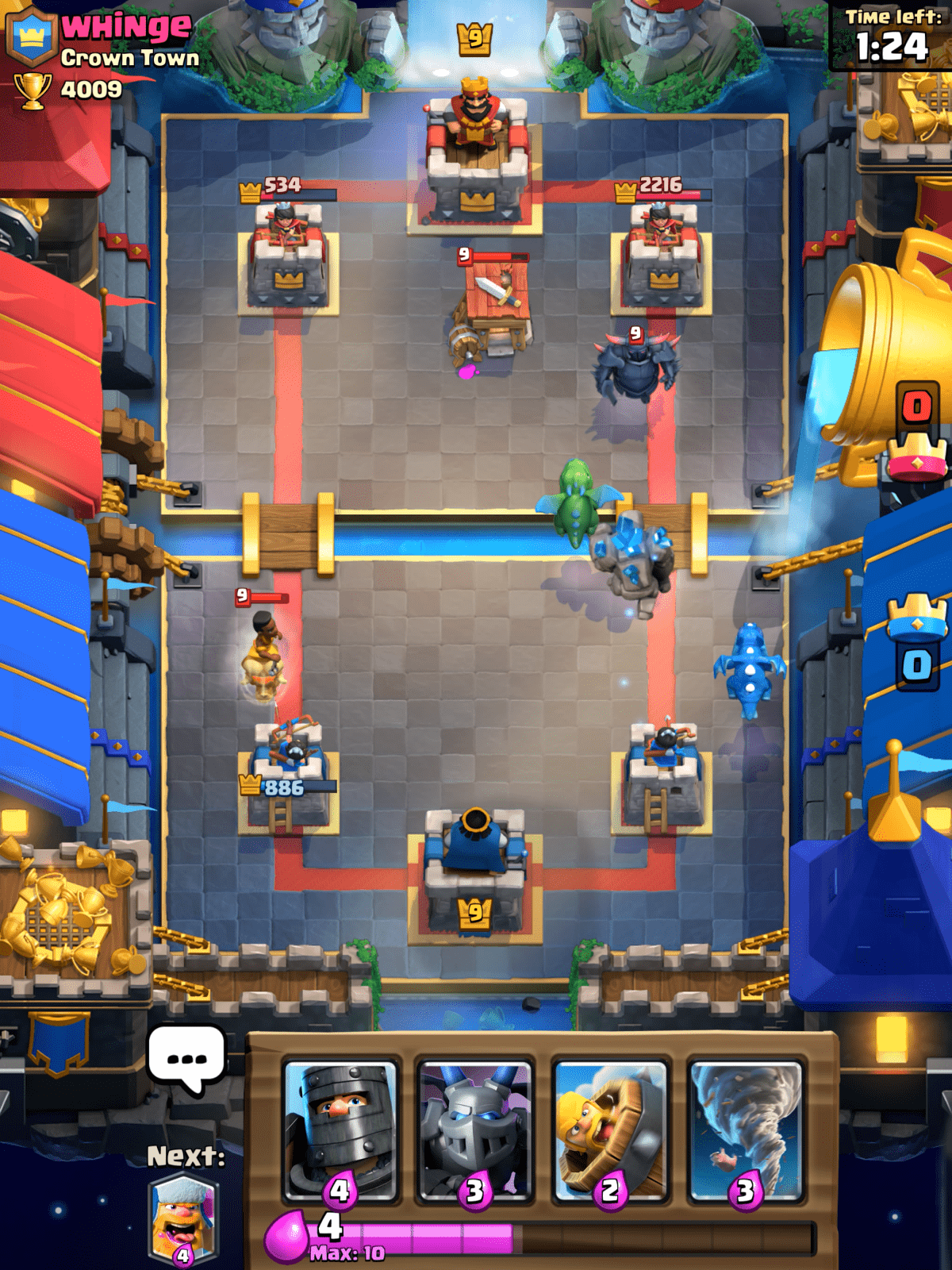}
\hfil
\includegraphics[height=50mm, trim=5 0 95 0, clip]{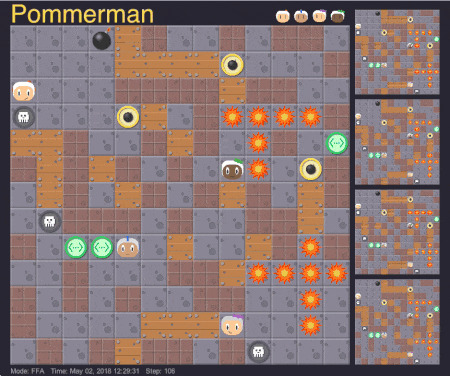}
\caption{\color{diffcolor} Screenshot of Clash Royale (left) and Pommerman (right).}
\label{fig:cr-screenshot}
\end{figure}

Battles in Clash Royale are played on a visually immersive $18~\times~32$ board initially consisting of a king tower and two princess towers for each player (see Fig.~\ref{fig:cr-screenshot}). The gameplay primarily consists of players deploying cards from their hand onto the battle arena to destroy the towers of the opponent.
Each card has an \emph{Elixir} cost associated with it and a card can only be deployed if the player has enough Elixir. 
Once a card is deployed in a specific location, it creates a troop or building or spell in the battle arena that follows predefined behaviours, and the player is dealt a new card from the deck. 
A battle ends instantaneously if a king tower is destroyed. If not, the player with the highest number of towers after three minutes wins.
Otherwise, the battle extends for an overtime of two minutes and the first player to destroy an enemy tower wins. 
Otherwise, the battle results in a draw. 

\begin{table}[t]
\caption{State space of a battle in Clash Royale}
\label{t:state}
\centering
\begin{tabular}{lll}
\toprule
& Feature & Description \\
\midrule
\multirow{5}{*}{Cards}
& Decks & \textnormal{Sets of $8$ cards chosen by both players for} \\
&  & \textnormal{this battle.} \\
& Level & \textnormal{Levels of all cards in the chosen decks.} \\
& Cost & \textnormal{Elixir cost of all cards in the chosen decks.} \\
& Hand & \textnormal{Sets of $4$ cards currently available to both} \\
& & \textnormal{players.} \\
& Next card & \textnormal{Next card to be dealt to each player.} \\
\midrule
\multirow{5}{*}{Progress}
& Elixir & \textnormal{Elixir currently available to each player.} \\
& Time & \textnormal{Time elapsed since beginning of battle.} \\
& 2x Elixir & \textnormal{Is battle in double elixir mode?} \\
& Overtime & \textnormal{Is battle in sudden death mode?} \\
& Past actions & \textnormal{List of actions by the player} \\
& & \textnormal{in the past 10 steps.} \\
\midrule
\multirow{5}{*}{Objects}
& Type & \textnormal{Types of all objects in the battle arena.} \\
& Position & \textnormal{$x$, $y$ coordinates of all objects in the arena.} \\
& Level & \textnormal{Levels of all objects in the arena.} \\
& Health & \textnormal{Health of all objects in the arena.}  \\
& Color & \textnormal{Object belongs to blue or red player?} \\
\bottomrule
\end{tabular}
\end{table}

\color{diffcolor}
The state $s$ of Clash Royale is comprehensively defined in Table~\ref{t:state}.
\color{black}
Each player observes the state of the battle arena, battle progress, the player's own hand and the next card. Information about the cards of the other player is not visible.
At any game state, player $i$ can choose either to deploy a legal card (a card that costs less than or equal to available elixir) or to wait for one time step. 
\color{diffcolor}
In this paper, an agent interacts with the Clash Royale game engine such that one time step corresponds to 0.5 seconds.
\color{black}
The action $a_i$ of deploying card $c$ by player $i$ can be represented as a tuple $(c, x, y)$ where $c$ is a card identifier and $(x, y)$ is the deploy position in the discrete $18 \times 32$ battle arena.
The action of waiting is represented with a special \emph{Wait} card.
Additionally, we augment the action space with cards in the hand that are illegal (with not enough Elixir). Choosing an illegal card forces the agent to intentionally wait until that card becomes available, after which it can choose to deploy any legal card or wait further.
The action space augmented in this way aids uniform exploration of all cards in the game and we use this in all our experiments.

Although the rules of Clash Royale are easy to learn, the game has great depth coming from predicting your opponent's moves, including their predictions of yours, which makes it hard to master.
Playing Clash Royale effectively requires a well coordinated combination of attacks and defenses and fast adaptation to the opponents' deck and style of play.
Further, because of limited Elixir resources and hidden information, waiting for a good deploy time is an important part of strategy.
Below, we describe the various scientific challenges in learning to play Clash Royale:
\begin{itemize}
\item \textbf{Cyclic strategies.} The cards in the game are designed such that each card can be countered effectively with another card (that is, Clash Royale is a \emph{non-transitive game}). Like the game of rock-paper-scissors, there is no single best deterministic strategy. 

\item \textbf{Partial observability.} Cards of the opponent are hidden and are only revealed throughout the opponent's deploys. 
Players can deceive their opponents by choosing to hide cards (not deploy) until later in the game (akin to bluffing).

\item \textbf{Exploration.} At any time during a battle in Clash Royale, only \emph{legal cards} (cards with costs less than the currently available Elixir) can be deployed. Naive exploration methods that choose random actions at each step leads to a greedy strategy of almost always deploying the card with the lowest cost (and thereby depleting Elixir). Good exploration strategies have to intentionally wait for the costlier cards.

\item \textbf{Dynamic, large, and discrete action space.} Clash Royale has a large discrete action space with the possibility to deploy any of 100 cards in the $18 \times 32$ arena ($\sim$60,000 discrete actions). However, at a particular time in a battle, it is only possible to deploy from the legal cards in the hand.
\end{itemize}

\subsection{Pommerman}

Pommerman is a popular multi-agent RL benchmark based on the classic Nintendo game Bomberman.
Battles in Pommerman are played on a $11 \times 11$ board initialized randomly with rigid walls and wooden walls (that may contain some power-ups) and four players near each corner (see Fig.~\ref{fig:cr-screenshot}).
The players can move in horizontal or vertical directions (that are not blocked by walls or bombs), collect power-ups or lay bombs in their current locations.
A player dies when they are on a tile affected by a bomb blast and effective gameplay requires strategic laying of bombs to knock down all of the opponents.
Hidden information in Pommerman consists of power-ups hidden inside wooden walls and the power-ups collected by other players.
The Pommerman benchmark consists of different scenarios and we consider the Free-For-All (FFA) variant in this paper. The goal of each agent in the FFA mode is to be the last agent to stay alive within a fixed-length episode of 800 timesteps.
The challenges in performing tree search on Pommerman involves: 1) the large branching factor (upto 1296) caused by four players simultaneously choosing from six actions, 2) the difficulty in credit assignment due to the presence of four players, and, 3) the common noisy rewards caused by suicides.
\color{diffcolor}
To assist learning, we mask out actions that immediately leads players into walls or flames (suicide).

We use a Cython implementation of the Pommerman environment based on \cite{matiisen2018pommerman}. For clarity of our experimental setup and ease of reproducibility, we open source the code for our Pommerman experiments here: \url{https://github.com/rinuboney/l2p-pommerman}.
\color{black}


\section{Oracle Planner With Full Observability}
\label{s:oracle}

In our approach, we first build an oracle planner which has access to the full game state. 
The goal of planning is to discover the optimal sequence of actions that maximize expected rewards. 
A dynamic programming approach to the planning problem involves estimating expected rewards for every legal action in each state, after which one can act greedily by choosing the action with the largest expected reward. 
A policy $\pi_i$ of player $i$ is a distribution over actions available in state $s$ for player $i$, that is, $a_i \sim \pi_i(a_i | s)$.
Let $\pi(a|s)=\pi_1(a_1|s)\pi_2(a_2|s)$ be the joint policy followed by players $i \in \{1, 2\}$.
Let $z \sim p(z|s, \pi)$ be the probability distribution over the set of all terminal states induced by following policy $\pi$ from state $s$.
The state value function $V_i(s)$ is the mean reward of player $i$ while players follow policy $\pi$ from state $s$:
\begin{equation}
V_i(s) = \E_{z \sim p(z|s, \pi)}[R_i(z)]
\label{eq:v}
\end{equation}
The state-action value function $Q_i(s, a)$ is the mean reward of player $i$ while players first take actions $a = (a_1, a_2)$ and then follow policy $\pi$ from state $s$:
\begin{align}
Q_i(s, a) = \E_{z \sim p(z|s, a, \pi)}[R_i(z)]
\end{align}
A possible way to do planning is to estimate $Q_i(s, a)$ for each player and choose the action for each player which maximises its expected reward. One problem with this approach is that one has to consider all combinations of actions $(a_1, a_2)$, which is prohibitive in games like Clash Royale where each player chooses from tens of thousands of actions.


In this paper, we take a different approach. We assume that the actions $a_1$ and $a_2$ are chosen independently, that is, we estimate $Q_i(s, a_i)$ taking an expectation over the opponent policy:
\begin{equation}
Q_i(s, a_i) = \E_{z \sim p(z|s, a_i, \pi)}[R_i(z)]
\,.
\label{eq:q_i}
\end{equation}
With this approximation, the problem formulation can be seen as a Partially Observable Markov Decision Process (POMDP) from the perspective of each player, where the opponent is subsumed into the stochastic environment.
At the end of planning, each player independently chooses the action that maximises the estimated Q values:
\[
a_i = \argmax_{a_i \in A_i(s)} Q_i(s, a_i)
\,.
\]


\subsection{Monte Carlo Search (MCS)}

Monte Carlo search (MCS) \cite{anthony2019policy} is a simple search method where $Q_i(s, a_i)$ is estimated for all actions $a_i \in A_i(s)$ by performing several iterations of random \emph{rollouts} from state $s$.
That is, both players estimate $Q_i(s, a_i)$ assuming that policies $\pi_1$ and $\pi_2$ are uniform distributions over the legal actions in every state.
In practice, we perform random rollouts for a fixed number of steps and then use a value function estimate $V$ to evaluate the final state.
In each iteration of MCS from state $s$, both players independently and randomly choose actions $a_i \in A_i(s)$ and continue to do so 
for a fixed number of steps (planning horizon), to reach state $\tilde{s}$.
At the end of an iteration, the estimate of $Q_i(s, a_i)$ is updated based on the 
value estimate $V(\tilde{s})$.


\color{diffcolor}

\subsection{Multi-Armed Bandits (MAB)}
\label{sec:mab}

Monte Carlo search can be improved by exploring more promising actions more often. This can be achieved by viewing action selection as a multi-armed bandit (MAB) problem: In the current state $s$, player $i$ has to choose an action $a_i \in A_i(s)$ with maximum expected reward. There are $|A_i(s)|$ arms and player $i$ can explore new actions or exploit actions with highest value estimates.
When MCS is enhanced by MAB, the MAB selection is done at the current state $s$ and the value estimates $Q_i(s, a_i)$ are obtained as in MCS by performing random rollouts.

In this paper, we use a \emph{decoupled} approach to action selection: each player independently chooses an action $a_i \in A_i(s)$ using its own instance of an MAB, thus the opponents are subsumed into the  stochastic  environment. We consider two popular MAB algorithms: the Upper Confidence Bound (UCB) and Thompson sampling. 

\subsubsection{Upper Confidence Bound}

UCB algorithms estimate the upper confidence bound that any given action is optimal \cite{browne2012survey, silver2017mastering}. While there exist different variations of UCB, we consider the commonly used UCB1 variant introduced in \cite{auer2002finite}. Each player $i$ independently estimates the upper confidence bound $\UCB_i(s, a_i)$ for each action $a_i \in A_i(s)$ as:
\begin{align}
\UCB_i(s, a_i) = Q_i(s, a_i) + c \sqrt{\frac{\log N}{n_{a_i}}} \,,
\label{eq:ucb}
\end{align}
where the $c$ hyperparameter controls the exploration-exploitation trade-off, $n_{a_i}$ is the visit counts of action $a_i$ and $N = \sum_{a_i \in A_i(s)} n_{a_i}$.

In each iteration, the action with the highest UCB value is chosen deterministically. At the end of planning, normalized visit counts define a probability distribution over actions. The final action can be chosen stochastically by sampling from this distribution of by deterministically choosing the action with the highest visit count.

\subsubsection{Thompson Sampling (TS)}

Thompson sampling \cite{thompson1933likelihood} maintains probability distributions of cumulative rewards for each action and chooses actions according to the probability that they are optimal.
Since the rewards in Clash Royale and Pommerman are binary,
the probability that taking action $a_i$ will lead to a win can be modeled using the Bernoulli distribution. 
The mean parameter $\theta_{a_i}$ of the Bernoulli distribution can be modeled with  a Beta distribution which is the conjugate prior distribution for the Bernoulli likelihood.
The parameters of the Beta ditribution can be updated by maintaining win and loss counts ($S_{a_i}$ and $F_{a_i}$ respectively) for each action.
Note that this posterior update assumes independent samples from a Bernoulli distribution, even though this is not true in a multi-agent setting.
During each iteration of planning, the action is chosen as
\begin{align*}
    a_i &= \argmax_{a_i \in A_i(s)} \theta_{a_i}
\\
   \theta_{a_i} &\sim \Beta(S_{a_i} + \alpha, F_{a_i} + \beta)
\end{align*}
In all the experiments in the paper, we set $\alpha=\beta=1$ and do not tune these hyperparameters. At the end of planning, the final action can be chosen stochastically in a similar manner or deterministically based on the estimated means of the Beta distributions.

\begin{figure*}[t]
\centering\scriptsize
\includegraphics[width=.8\linewidth]{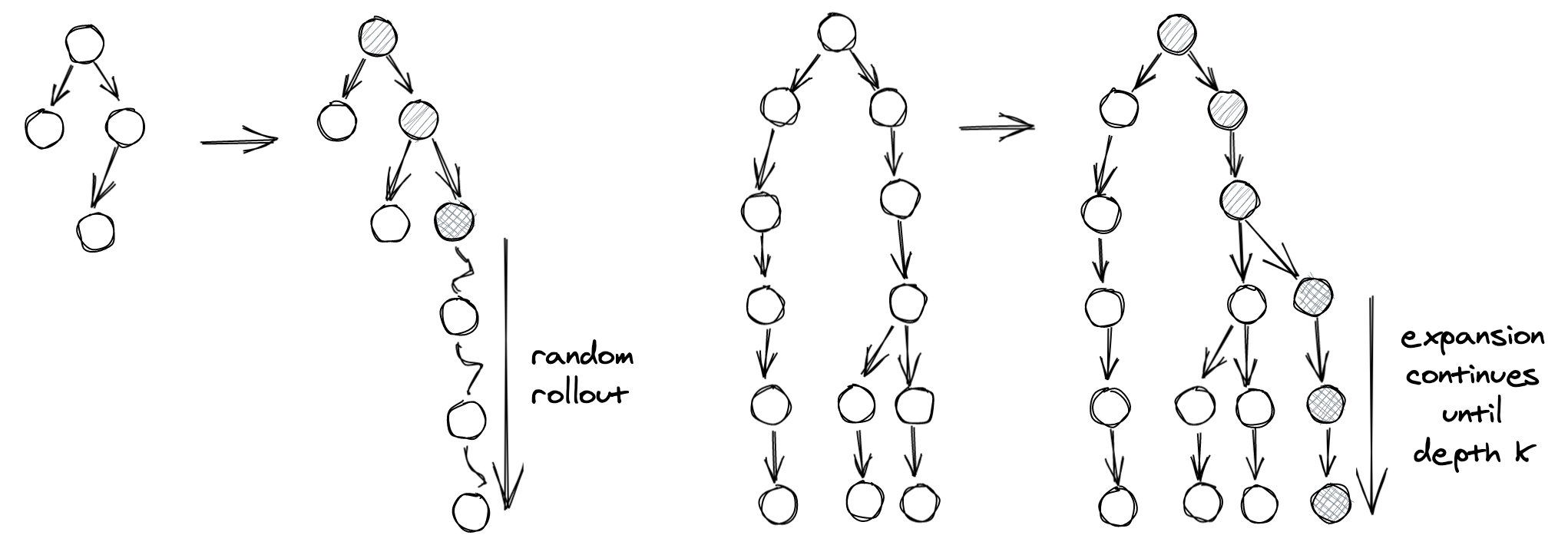}
\\
\hspace{-27mm}MCTS\hspace{65mm}FDTS
\caption{\color{diffcolor} Illustration of how a search tree is modified in one planning iteration in MCTS (left) and FDTS (right).}
\label{fig:fdts}
\end{figure*}

\subsection{Monte Carlo Tree Search (MCTS)}

MCS described previously has several limitations: 1) it only plans actions for the current state and hence cannot discover effective action combinations, 2) it discards all information about future states and actions traversed during rollouts and plans from scratch in each step, and 3) the rollout policy is random and hence the estimated 
$Q$ values are under the assumption that both players will act randomly in the future. 

MCTS builds upon MCS by considering action selection in all states encountered during rollouts as a multi-armed bandit (MAB) problem. MCTS is a best-first tree search algorithm and begins from a root node corresponding to current state $s$. We start with the most
common variant of MCTS in which each MCTS iteration from current state $s$ consists of the following steps:
\begin{enumerate}

\item \textbf{Selection-expansion.} Starting at the root node (which corresponds to the current state of the game), a \emph{tree policy} is used to descend through the tree  until a new state $s'$ is reached. In the case of two players acting simultaneously, the tree policy can be implemented by both players independently choosing actions $a_i \in A_i(s)$ using one of the MAB algorithms discussed in Section~\ref{sec:mab}.


\item \textbf{Evaluation.} The value $V(s')$ of the new state $s'$ is evaluated, which can be done in different ways: 1)~by applying a handcrafted or a learned value function to $s'$, 2)~by random rollout(s) from state $s'$ until a terminal state $z$ and using
$R(z)$ as a Monte Carlo estimate of the value, or 3)~by a fixed length rollout and applying a value function to the reached state.

\item \textbf{Backup.} The values $Q_i(s, a_i)$ for all the ancestors of node $s'$ are updated using the estimate $V(s')$ and the visit counts $n_{a_i}$ are incremented by one.

\end{enumerate}
See Fig.~\ref{fig:fdts} for a simplified illustration of one MCTS iteration.

After several planning iterations, both players independently choose their best actions and the search tree built by MCTS is re-used for planning in subsequent states by moving the root node to the child node corresponding to the chosen joint action. MCTS allows for discovery of effective sequence of actions, reuse of statistics computed from previous states and iterative improvement of the rollout policy.

A potential problem with MCTS is that the selection-expansion step may stop very early in the tree. This is likely to happen in the games with a large branching factor of the search tree.
It is very probable that the tree policy will encounter a novel game state in one of the upper levels of the tree, after which the state is evaluated. This can limit the effective planning horizon of MCTS and makes it problematic to properly evaluate long-term plans.


\begin{figure*}
\centering
\includegraphics[width=\linewidth, trim=0 12 0 12, clip]{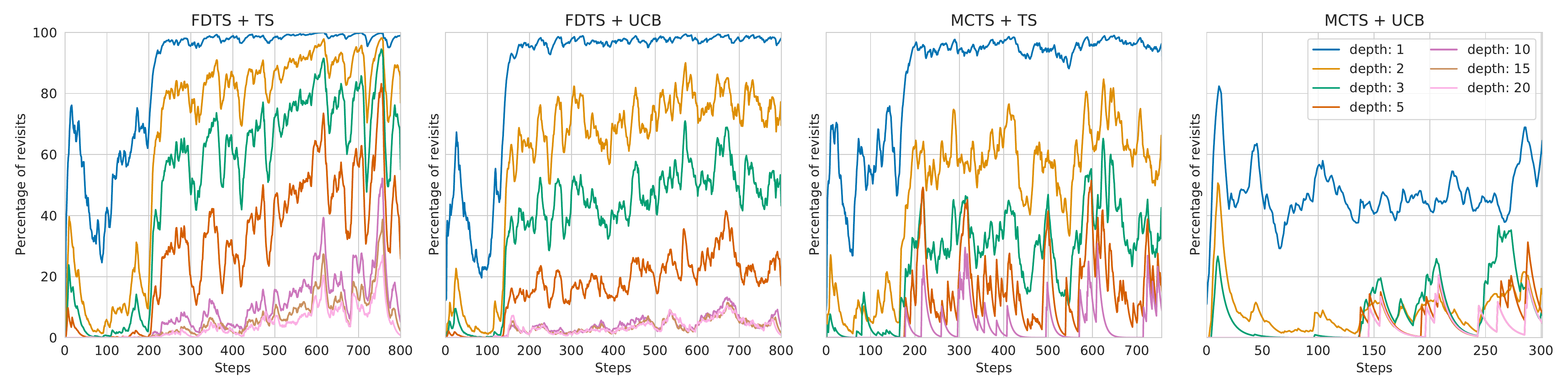}
\caption{\color{diffcolor}
Comparison of different planners based on the reuse of information stored in the search tree, in a game of Pommerman. A Pommerman game can last for a maximum of 800 steps and in each step we execute the planning procedure for 100 iterations (and a fixed horizon of 20 in the case of FDTS). We plot the (low-pass filtered) ratio of state revisits during planning at each game step (that is, in all of the times the planner visits a state in depth $d$ of the search tree, the ratio of states that it has previously visited.)
We use this to measure the effectiveness of use of information stored in the search tree.
The best-performing FDTS+TS planner frequently reuses information, even up to the maximum depth of 20.
\color{black}
}
\label{fig:revisits}
\end{figure*}

\subsection{Fixed-Depth Tree Search (FDTS)}

We propose to improve MCTS by encouraging planning at least several steps ahead from the current state. The proposed algorithm, that we call \emph{fixed-depth tree search (FDTS)}, consists of the following steps:
\begin{enumerate}

\item \textbf{Selection-expansion-rollout.} Starting at the root node, an MAB tree policy is applied exactly $k$ times to descend through the tree. If the game reaches a novel state at a particular level, a new node is added to the tree and the tree policy continues action selection from that node until a desired depth level $k$ is reached. This steps results in creating a new branch with a leaf node with state $s'$ at a particular depth level.

\item \textbf{Evaluation.} The value of the node state $s'$ reached at depth $k$ is evaluated. In our experiments, the evaluation step is done by applying a handcrafted value function without performing random rollouts.

\item \textbf{Backup.} The values $Q_i(s, a_i)$ for all the ancestors of node $s'$ are updated using the estimate $V(s')$.

\end{enumerate}
One iteration of FDTS is illustrated in Fig.~\ref{fig:fdts} and the Python pseudocode for FDTS can be found in Listing~\ref{code:fdts}.

The proposed algorithm can be viewed as combining in one step the selection-expansion step and the fixed-length rollout part of the evaluation step of classical MCTS.
After a novel state is reached, the MAB algorithm is recursively used to expand that node into a branch that reaches a fixed tree depth $k$. This is essentially equivalent to a random rollout. The important difference is that we add nodes to the tree for all the states encountered during the random rollout.

Keeping the trajectories encountered during random rollouts may seem wasteful, especially for problems with a large branching factor. However, this turns out to work well in the games considered in this paper because the MAB selection process systematically re-visits nodes existing in the tree despite of the large branching factor.
In Fig.~\ref{fig:revisits}, we demonstrate that the FDTS equipped with UCB and especially with TS re-uses information collected in the previous planning steps. The increased percentage of re-visited nodes in FDTS compared to MCTS suggests that storing the rollout trajectories in the search tree is indeed beneficial. The same figure shows that Thompson sampling tends to re-visit existing nodes more often than UCB and this further improves the quality of planning, which is supported by our experimental results.

\subsection{Memory and Computation Requirements}

MCS is simple to implement and has minimal memory and computation requirements. MCS only stores statistics of legal actions in the current state. MCTS and FDTS requires storing statistics of legal actions in all previously visited states of an episode. The main computation in MCS is the stepping forward of the game state using the game engine. MCTS and FDTS further requires more computation at every state for action selection using an MAB algorithm. 






\color{black}

\section{Experiments on Planning With the Oracle}

In this section, we evaluate the proposed planning algorithms on the games of Pommerman and Clash Royale. Although optimal policies in multiplayer games are stochastic, similar to \cite{perick2012comparison}, we observe that deterministic policies perform better in practice. In all the experiments presented in this paper, we deterministically choose the action with the highest value at the end of planning.

\subsection{Pommerman}

Since Pommerman FFA is a four player game, we compare different planning algorithms by pitting them against three copies of the strong rule-based agent that is provided along with the Pommerman environment.
\color{diffcolor}
It is important to note that the proposed algorithms perform planning in the self-play mode using decoupled action selection for each player, that is they are not aware of the policy of the rule-based agents. Planning against known agents would be a much easier task. 

In Pommerman, the number of legal actions for each player can vary from 1 to 6, that is, the branching factor of the search tree can vary from 1 to 1296. 
In all our experiments, we perform 100 simulations of the planning algorithm at every time-step and use a planning depth of $k=20$ (in MCS and FDTS).
\color{black}
In the evaluation step of tree search, we simply use the reward function of the Pommerman as the value function \cite{matiisen2018pommerman}.

\color{diffcolor}
In Table~\ref{t:pommerman_oracle}, we report the number of wins, draws and losses in 400 games for different settings. We consider three planning algorithms: MCS, MCTS and FDTS, and two alternative ways for actions selection: Thompson sampling and UCB1 with $c=2$.
For a fair comparison to MCS and FDTS, we use MCTS with random rollouts (at the end of the expansion step in an MCTS iteration, we perform random rollouts till a fixed depth  of 20 and use that state for evaluation), which is similar to FDTS except that we do not add the nodes visited during the random rollouts to the search tree.
A comparison of MCTS performance with and without this random rollouts is reported in Table~\ref{t:mcts-random}.
The best results are obtained with FDTS+TS which attains a win-rate of $51.3\%$ with no reward shaping. A similar setup of self-play planning on a Java implementation of the Pommerman environment was considered in \cite{perez2019analysis} who reported win rates of $46.5\%$ for MCTS and $33.0\%$ for Rolling Horizon Evolutionary Algorithm \cite{perez2013rolling} using shaped rewards. 
\color{black}


\begin{table}[!t]
\caption{A four-player Pommerman FFA: Percentages of wins, draws and losses against three rule-based opponents computed after 400 games. The first column is evaluation of the rule-based agent against three copies of itself.
}
\label{t:pommerman_oracle}
\centering
\begin{tabular}{l|c|cc|cc|cc} 
\toprule
 & \multirow{2}{*}{Rule-based} & \multicolumn{2}{c|}{MCS} & \multicolumn{2}{c|}{\color{diffcolor} MCTS} & \multicolumn{2}{c}{\textbf{FDTS}} \\ 
 & & UCB & TS & {\color{diffcolor} UCB} & {\color{diffcolor} TS} & UCB & \textbf{TS} \\ 
\midrule
Wins   & $19.2$ & $35.8$ & $36.3$ & {\color{diffcolor} $37.3$} & {\color{diffcolor} $40.8$} & $42.3$ & $\mathbf{51.3}$ \\ 
Draws  & $23.4$ & $41.0$ & $37.5$ & {\color{diffcolor} $32.4$} & {\color{diffcolor} $36.1$} & $32.5$ & $\mathbf{29.0}$ \\ 
Losses & $57.4$ & $23.2$ & $26.2$ & {\color{diffcolor} $30.3$} & {\color{diffcolor} $23.1$} & $25.2$ & $\mathbf{18.2}$ \\ 
\bottomrule
\end{tabular}
\end{table}

\subsection{Clash Royale}

In Clash Royale, the number of discrete actions $A_i(s)$ is very large but the actions are correlated: deploying a card on nearby positions tend to produce the same outcome. To approximate a good policy, we sample a random set of 64 positions from the space of legal positions for every legal card. A sufficiently large random set would include the optimal deploy positions. 
\color{diffcolor}
With this approximation, in Clash Royale, there are two players and the legal actions for each player (with the random sampling of deploy positions) can vary from 1 to 257. That is, the branching factor of the search tree can vary from 1 to 66049. \color{black}

In our experiments, we use simple handcrafted value functions
for oracle planning: we compute $V(s)$ by doing a rollout from state $s$ assuming that both players do not deploy any more cards. Since the consequences of already deployed cards 
have predefined behaviour, we can reach state $s'$ where the battle arena only contains towers. Then, we evaluate $V(s)$ using the terminal reward function $R(s')$.




We compare UCB1 with $c=1$, Thompson sampling and simple random sampling using Monte Carlo search, by pitting one MAB algorithm against another.
For example, to compare Thompson sampling with UCB, Player~1 performs planning using Thompson sampling for action selection of both players and Player~2 independently performs planning using UCB for action selection of both players. We compute the win rate of of an algorithm against another for 400 games in this setting.
The results are shown in Table~\ref{t:cr-mab}. Both Thompson sampling and UCB clearly outperform random sampling. Thompson sampling performs the best of all.

\color{diffcolor}
While UCB could potentially be fine-tuned to work better with a more comprehensive search over the hyperparameters or using a different UCB variants such as UCB1-Tuned \cite{auer2002finite}, we found Thompson sampling to robustly work well in most settings. 
To test the robustness of Thompson sampling, we compare it against UCB with different value of exploration hyperparameter $c$ and planning horizon/depth. The win rates of the comparison in Clash Royale is shown in Table~\ref{t:ucb}, where Thompson sampling clearly outperforms UCB in most tested settings.
\color{black}
We therefore use Thompson sampling as the MAB algorithm in all our further experiments.

\begin{table}[t]
\caption{Clash Royale: Comparison of MCS, MCTS and FDTS.
The planning horizon is $k=50$.
Shown are win rates and $95\%$ confidence intervals.
}
\label{t:cr-oracle}
\centering
\begin{tabular}{lc}
\toprule
 & Win rate \\
\midrule
MCTS vs MCS & $41.3\pm4.5\%$ \\
\textbf{FDTS} vs MCTS & $96.5\pm1.8\%$ \\
\textbf{FDTS} vs MCS & $80.3\pm3.9\%$\\
\bottomrule
\end{tabular}
\end{table}

\begin{table}[t]
\caption{Clash Royale: Comparison of MCS with MCTS and FDTS based on their win rates by evaluating each pair on 40 games. 
}
\label{t:mcts-ablation}
\centering
\begin{tabular}{ccc}
\toprule
Horizon & MCTS vs MCS &\textbf{FDTS} vs MCS \\
\midrule
10 & $43.7\%$ & $\mathbf{68.1\%}$ \\
25 & $19.4\%$ & $\mathbf{70.5\%}$ \\
50 & $41.3\%$ & $\mathbf{80.3\%}$ \\
\bottomrule
\end{tabular}
\end{table}


We compare MCS, MCTS and FDTS in Clash Royale by pitting one algorithm against another for 400 games, where each player independently performs planning using the assigned algorithm.
The results of our experiments are shown in Table~\ref{t:cr-oracle}. The proposed FDTS planning achieves the best performance.

For further comparison of MCTS and MCS, we pit the two variations of MCTS against MCS for different planning horizons. The win rates on 40 games of Clash Royale are shown in Table~\ref{t:mcts-ablation}. FDTS outperforms MCS on all planning horizons, with an increased difference for deeper search.
These results suggest that FDTS is able to discover better combinations of actions and re-uses statistical information (as demonstrated in Fig.~\ref{fig:revisits}) to outperform MCS for all planning horizons, with an improved performance as the planning horizon increases.

\color{black}


\section{Training Follower Policy with Partial Observability}
\label{s:follower}

Planning enables competitive play with generalization to unseen states. However, the oracle planner has two limitations: 1)~It performs many rollouts to make decisions in every state, requiring a game implementation that must run much faster than real-time, to be able to act in a real-time battle. 2)~The oracle planner cheats by having access to the full game state: private information like the deck and hand of the opponent in Clash Royale and hidden power-ups in Pommerman becomes visible during future states of planning rollouts. This could be avoided by randomizing hidden information during planning but the game engines of these games do not support this.

In our approach, we propose to use imitation learning to train a \emph{follower} policy network to perform similarly to the oracle planner but under real-time computation and partial observability. One straightforward way of doing this would be via cloning of the oracle behavior: one can
collect trajectories generated by the oracle planner with self-play and use that data to train the follower policy. However, this approach results in a relatively poor performance (see Table~\ref{t:pommerman-follower}).


\color{diffcolor}
We instead use the DAgger algorithm \cite{ross2011reduction} for better performance. In DAgger, the follower policy makes decisions during self-play and the oracle planner is used to compute better actions in the states encountered by the follower.
Since the follower is initially random, self-play encounters diverse states and the oracle planner provides stable training targets, leading to an efficient improvement of the follower.
The algorithm for training follower networks is listed in Algorithm~\ref{alg:main}.
\color{black}

\begin{algorithm}[H]
\caption{Learning to play imperfect-information games by imitating an oracle planner with DAgger}
\begin{algorithmic}[1]
\STATE Initialize follower policy $\pi_f$ and replay buffer $\mathbb{D}$.
\FOR{each episode}
\STATE Initialize oracle planner $\pi_o$.
\FOR{time $t$ until the episode is over}
\color{diffcolor}
\STATE Compute follower actions $a_f = \pi_f(o_t)$ from partial observations $o_t$, and apply them to the game.
\STATE Compute oracle actions $a_o \sim \pi_o(s_t)$ using (self-play) tree search, with access to full state $s_t$.
\STATE Add data $(o_t, a_o)$ to replay buffer $\mathbb{D}$ and train follower policy $\pi_f$ using $\mathbb{D}$ to predict the oracle actions $a_o$ from partial observations $o_t$.
\color{black}
\ENDFOR
\ENDFOR
\end{algorithmic}
\label{alg:main}
\end{algorithm}


\section{Experiments on Training the Follower}

We train follower networks to imitate the oracle planner by predicting the oracle action from partial observations $o_i$.
The oracle is chosen to be the best performing fixed-depth tree search (FDTS) with Thompson sampling (TS). 

\subsection{Pommerman}

In Pommerman, 
we train a follower network to imitate the oracle planner on 500 battles. 
We use the same network architecture as \cite{DBLP:journals/corr/abs-1809-07124}.
The observations are represented in a $11 \times 11$ spatial representation (corresponding to the $11\times11$ board in the game), with 14 feature maps. The features represent presence and positions of 10 different objects in the board, bomb blast positions and lifetime and the power-ups collected by the agent. The network architecture consists of four convolutional layers with 32 channels (with ReLU activations) and a final linear layer that predicts the softmax probabilities of the six discrete actions.
We used random search to tune the hyperparameters of the oracle planner and the follower policy. All hyperparameters along with their search range and final values for Pommerman are reported in Table~\ref{t:pommerman-hp}.

\begin{table}[!t]
\caption{A four player Pommerman FFA: Comparison of follower agents
against three rule-based baseline opponents on 400 games.
Evaluation of the rule-based agent against three copies of itself is reported in the first column for reference.
}
\label{t:pommerman-follower}
\centering
\begin{tabular}{l|c|cc}
\toprule
 & \multirow{2}{*}{Rule-based} & \multicolumn{2}{c}{\textbf{Follower}} \\
 & & \textbf{DAgger} & Oracle-behavioral cloning \\
\midrule
Wins   & $19.2\%$ & $\mathbf{23.3\%}$ & $17.4\%$\\
Draws  & $23.4\%$ & $\mathbf{22.5\%}$ & $19.2\%$ \\
Losses & $57.4\%$ & $\mathbf{54.2\%}$ & $63.4\%$ \\
\bottomrule
\end{tabular}
\end{table}

\begin{table*}[t]
\caption{Clash Royale: Comparison of all agents based on win rates (and $95\%$ confidence intervals) of each pair evaluated on 100 games. Win rates of Q-MC and Follower are averaged over win rates of networks trained used 5 different seeds.}
\label{t:cr-follower}
\centering
\begin{tabular}{lccccc} 
\toprule
 & Random & Q-MC & Human-BC & Follower & Oracle \\ 
\midrule
Random              & -                         & $20.9\pm7.9\%$            & $1.0\pm1.9\%$             & $1.4\pm2.3\%$     & $0.0\pm0.0\%$          \\ 
Q-MC                 & $79.1\pm7.9\%$            & -                         & $14.9\pm6.9\%$            & $8.6\pm4.3\%$    & $0.0\pm0.0\%$          \\ 
Human-BC            & $99.0\pm1.9\%$            & $85.1\pm6.9\%$            & -                         & $28.6\pm8.8\%$    & $6.1\pm4.7\%$          \\ 
\textbf{Follower}   & $\mathbf{98.6\pm2.3\%}$   & $\mathbf{91.4\pm4.3\%}$   & $\mathbf{71.4\pm8.8\%}$   & -                 & $\mathbf{6.6\pm4.8\%}$ \\ 
Oracle              & $100.0\pm0.0\%$           & $100.0\pm0.0\%$           & $93.9\pm4.7\%$            & $93.4\pm4.8\%$    & -                      \\ 
\bottomrule
\end{tabular}
\end{table*}

We evaluate the Pommerman follower against three rule-based opponents and the results are shown in Table~\ref{t:pommerman-follower}. The follower agent trained with DAgger is able to achieve a win rate 23.3\%, outperforming the rule-based agent and even the MCTS oracle planner. Note that previous works have achieved high win-rates against the rule-based opponent by directly training against it \cite{matiisen2018pommerman, gao2019skynet}. Instead, we learn purely from self-play, which yields an agent that is able to compete with different kinds of opponents.

\subsection{Clash Royale}

In Clash Royale, we train a follower network to imitate the oracle planner on 300 battles. The follower is a convolutional neural network that predicts the $Q$ values (means of the Beta distributions)
of the spatial deploy positions for all legal cards.
During self-play and evaluation, the follower network deterministically chooses the action with the largest $Q$ value.
In Clash Royale, the objects in the battle arena, battle progress, current cards and the past 10 actions are represented using $18\times32$ spatial feature maps (corresponding to the $18\times32$ battle arena in the game). Card types and object types in the battle arena are represented using learnable embeddings.
The follower network predicts the $Q$ values of all deploy positions of all legal cards based in these spatial features. 
All hyperparameters of the follower network along with the random search range and final values are reported in Table~\ref{t:cr-hp}.

We evaluate the follower against three baseline agents: 1) \emph{Random}: a simple uniform random policy, 2) \emph{Q-MC}: a model-free agent trained with Monte Carlo value targets \cite{adkb2018td}, and 3)~\emph{Human-BC}: a strong agent trained to imitate human actions. 
DQN \cite{mnih2015human} was not included in the comparison because it was unstable, most likely due to the large action space and delayed actions.

\color{diffcolor}
\textsc{Human-BC} is a very strong baseline: it is a mature agent that has been in production for over a year.
That agent was trained using behavioural cloning (supervised learning) to imitate human actions from 76 million frames of human replay data from Clash Royale. These replays consisted of games played by humans with a good skill level, all from 4000 trophies and above, and played with a diverse set of decks. The architecture of \textsc{Human-BC} and the training parameters were tuned for metrics like prediction accuracy of deployed cards and their deploy positions. 
The \textsc{Human-BC} agent consists of two feature extraction networks and an action prediction network. A battle arena feature extraction network embeds the objects (along with their features) in the battle arena in a spatial grid 
based on their positions and extracts features from the spatial inputs using residual blocks. A battle context feature extraction network extracts battle context features based on cards and battle progress, similar to the follower network architecture, but with a larger network consisting of residual blocks. The battle arena and battle context features are combined using a sum operation and an action prediction network consisting of residual blocks predicts: 1)~when to deploy, 2)~card to be deployed, 3)~deploy position, and 4)~value of current state (auxiliary task). The predicted card is deployed onto the predicted deploy position only if the policy predicts that it should be deployed in the current step.

The win rates of all pairs of agents 
are presented in Table~\ref{t:cr-follower}.
The Q-MC agent does not perform very well as it is able to beat only the random agent. By analyzing its playing style, one can notice that it tends to learn a particular strategy that is easily predictable by human players. 
The Human-BC agent is very competitive, the analysis of its gameplays suggests that is able to use strategies which are common for human players.
\color{black}

The oracle planner beats the other agents almost always, which is natural because it has access to more information.
By analyzing its gameplays, we observed that the oracle planner was able to discover effective strategies commonly used by human players.\footnote{\color{diffcolor} See \url{https://sites.google.com/view/l2p-clash-royale} for our supplementary video including the gameplay videos.} Some of the discovered strategies are
\begin{enumerate}
\item \textbf{Groups of troops.} 
The planner is consistently playing high-hitpoint ``tank'' troops like \emph{Giants}, \emph{Knights}, or \emph{Baby Dragons} in the front, and support units like \emph{Musketeers} or \emph{Archers} behind the tank. This is a key strategy for successful attacks that requires coordinating deploys across several timesteps.

\item \textbf{Defense against tanks.} When attacked by a single tank unit without support units, the planner deploys high DPS (damage per second) troops like \emph{Musketeer} or \emph{Minions} to directly and efficiently remove the tank. However, if there are support units behind the tank, then the defending planner typically tries to destroy the support units first, to minimize potential tower damage from such more threatening attacks.

\item \textbf{Hedging.} Clash Royale games often have pivotal moments where one of the players must decide between two high level strategies: trying to defend against an oncoming attack, or hedging bets by skipping defense and launching a similarly powerful attack on the other lane. The planning agent is able to decide to forgo defense and respond with an attack against the other tower.

\item \textbf{Slowing down attacks.} If an attack is approaching but there are no good defense cards in the hand, the planner is able to deflect a threatening attack by deploying a tank like \emph{Giant} to slow down the attack and thus rotating more suitable cards to the hand.

\item \textbf{Race against time.} In the end of the game, when both players are equally close to winning, it's essential to damage the opponent's king tower quicker than the opponent damages yours. In these scenarios, the planner is coordinating all deploys at the king tower, using even weak damage from spells like \emph{Arrows}.
\end{enumerate}

\color{diffcolor}
Training the follower network with the oracle supervision resulted in a Follower agent which outperforms the very strong \textsc{Human-BC} baseline. Although the Follower does not have access to the full game state, it successfully uses the strategies discovered by the oracle, which we observed by analyzing its playing style.
\color{black}


\section{Related Work}

Previous works on RL in games with high-dimensional state-action spaces such as StarCraft~II \cite{vinyals2019grandmaster}, Dota~2 \cite{berner2019dota} and Honor of Kings \cite{ye2019mastering} have used model-free RL algorithms \cite{schulman2017proximal, espeholt2018impala}, requiring a large amount of data to learn. We take a model-based planning approach to learn to play imperfect-information games. 
Previous works have found MCTS to be an effective planning algorithm in various simultaneous-move games with low-dimensional state-action spaces \cite{shafieicomparing, teytaud2011lemmas, bovsansky2016algorithms}, even though it does not have any theoretical guarantees on achieving optimal play in simultaneous-move or imperfect-information games and can be exploited by regret minimization algorithms \cite{shafieicomparing}.
MCTS has been used for planning in imperfect-information games essentially by \emph{determinization} of the hidden information \cite{frank1998search, ginsberg2001gib, bjarnason2009lower}, also known as Perfect Information Monte Carlo (PIMC) \cite{long2010understanding}. The determinization technique involves performing several instances of the MCTS procedure with different randomizations of the hidden information and average across the resulting policies. Information Set MCTS (IS-MCTS) \cite{cowling2012information} involves determinization of hidden information in each MCTS iterations to construct a search tree of information sets. 
MCTS algorithms that use determinization \cite{frank1998search, ginsberg2001gib, bjarnason2009lower, cowling2012information} are not applicable to complex games or real-world problems, where it is not possible to randomize hidden information. 
In this paper, we introduce an algorithm for efficient planning and learning in imperfect-information games by using a function approximator to average across the resulting policies produced by an oracle planner that has access to the hidden information.
Even though averaging across different actions computed by the oracle in different states are not optimal,
similar to previous works \cite{frank1998search, ginsberg2001gib, schafer2008uct, buro2009improving, bjarnason2009lower}, we found it effective in learning strong policies.

Learning to play card-based real-time strategy (RTS) games was previously considered in \cite{liu2019} using DQN to learn to select cards and computing the deploy positions in a post-hoc manner using an attention mechanism, which is suboptimal as the deploy positions are never trained.

Guo et al. \cite{guo2014deep} used imitation learning of an MCTS planner in the simpler single-player setting of Atari games, with full obervability and a small number of discrete actions. We show that the naive MCTS used in \cite{guo2014deep} is problematic in imperfect-information simultaneous-move games with large action spaces and introduce fixed-depth tree search with Thompson sampling for better planning.

\color{diffcolor}
Combinatorial multi-armed bandit (CMAB) algorithms can be applied in settings where the action space of each player consists of combinations of multiple variables \cite{gai2010learning, chen2013combinatorial, ontanon2013combinatorial}. For example, in Clash Royale, an action consists of a card and the ($x$ and $y$) deploy position of the card. In this work, we resort to use of MAB algorithms as the combinations of 4 cards and a random sample of 64 deploy positions are limited to only 256 arms. Alternatively, CMAB algorithms can be used for a proper treatment of combinatorial action spaces with very large branching factors \cite{ontanon2017combinatorial}.
\color{black}

\section{Discussion}

\color{diffcolor}
We demonstrate good performance on learning to play in the novel setting of Clash Royale and the challenging multi-agent RL benchmark of Pommerman.
Our approach consists of an oracle planner that has access to the full state of the environment and a follower agent which is trained to play the imperfect-information game by imitating the actions of the oracle from partial observations.
We demonstrate that naive MCTS is problematic in high-dimensional action spaces. 
We show that fixed-depth tree search (FDTS) and Thomspon sampling overcome these problems to discover efficient playing strategies in Clash Royale and Pommerman.
The follower policy learns to implement them from scratch by training on a handful of battles. 
Our two-step approach can be combined in an iterative fashion by improving the oracle planner using $Q$ estimates from the follower policy. Potential directions of future work include exploration of regret minimization algorithms used in Poker \cite{zinkevich2008regret, moravvcik2017deepstack}.
\color{black}

While Clash Royale serves as a novel setting of reinforcement learning research, learned agents also have several use cases in game design. For example:
1)~agents can do automated testing of new game content, such as new cards or levels,
2)~agents can be used as practice opponents,
3)~new single player games can be designed where humans play against computer agents, and
4)~agents can provide assistance to new players during tutorial or unlocking of new cards.

\section{Societal Impacts}

The research presented in this paper can have 
an impact on the gaming industry. On the positive side, self-play algorithms can replace handcrafted rules which are currently most widely used for: 1)~designing bots that play a game in the place of a human, 2)~producing game content like \emph{boss levels} (fights against a strong computer-controlled enemy). Designing rule-based bots which are game-specific and difficult to maintain is an expensive component of game development, replacing this component with a general self-play algorithm can have strong impact on the industry. Self-play bots can also be easily retrained and used to reduce manual work for game testing, which involves finding bugs and assessing the difficulty levels of a game.
On the negative side, in the wrong hands, skillful bots can be used for cheating in the game, which is a major issue in video games, especially in online games \cite{blackburn2014cheating, zuo2016bad, paay2018motivations}. Bots can be used to cheat by providing unfair advantage to a player during gameplay. If players cannot know for sure that they are playing against other human opponents on equal grounding, it can erode the trust of the player community towards the game as a whole.
Similarly to any other RL algorithm, our research results alone are not enough to enable cheating in games in general, because the model would have to be first trained against a specific game environment, and then integrated to the game software, both of which require low level access to the game engine. Overall, further research in data-efficient RL will increase the risk of bot misuse in games, but dealing with that is a line of future work.

\color{diffcolor}
\appendix[]

\begin{lstlisting}[language=Python, caption=Fixed depth tree search, label=code:fdts]
def fdts(root_state, search_tree, n_simulations, fixed_depth):
    """ 
    Perform FDTS from root_state for n_simulations rollouts of length fixed_depth.
    search_tree is a dictionary that maps previously visited states to independent instances of an MAB algorithm for each player.
    """
    if root_state not in search_tree:
        search_tree[root_state] = [
            MAB(state.legal_actions(player))
            for player in state.players()
        ]
        # MAB is an implementation of a multi-armed bandit algorithm like UCB or Thompson sampling
    for _ in range(n_simulations):
        state = root_state
        search_path = []
        for _ in range(plan_horizon):
            decoupled_mab = search_tree[state]
            # Independently select actions for each player using the MAB algorithm
            actions = [
                mab.select() for mab in decoupled_mab
            ]
            state = state.apply(actions)
            search_path.append(decoupled_mab)

            if state not in search_tree:
                search_tree[state] = [
                    MAB(state.legal_actions(player))
                    for player in state.players()
                ]
                # Setting plan_horizon to a large value and adding a break statement here would result in the standard MCTS algorithm

            if state.is_terminal():
                break

        # Evaluate the final state using a handcrafted value function
        values = value_fn(state)

        # Update the statistics of all MAB instances visited during this rollout
        for decoupled_mab in search_path:
            for mab, value in zip(decoupled_mab, values):
                mab.update(value)

    # Independently select the final actions for each player
    actions = [
        mab.act() for mab in search_tree[root_state]
    ]
    return actions
\end{lstlisting}
\color{black}

\begin{table}[t]
\caption{Clash Royale: Comparison of MCS agents equipped with three MAB algorithms for action selection. The planning horizon is $k=50$. Shown are win rates and $95\%$ confidence intervals.}
\label{t:cr-mab}
\centering
\begin{tabular}{lc}
\toprule
 & Win rate \\
\midrule
UCB vs Random & $93.5\pm2.4\%$ \\
\textbf{TS} vs Random & $98.5\pm1.2\%$ \\
\textbf{TS} vs UCB & $64.0\pm4.7\%$\\
\bottomrule
\end{tabular}
\end{table}

\begin{table}[t]
\caption{Clash Royale: Win rates of Thompson sampling against UCB for different planning horizons and UCB exploration hyperparameter $c$. Each pair is evaluated on 50 games. 
}
\label{t:ucb}
\centering
\begin{tabular}{cccccc}
\toprule
Horizon & c = 0.5 & c = 1 & c = 2 & c = 3 & c = 4 \\
\midrule
10 & $64.0$ & $69.0$ & $77.6$ & $82.2$ & $82.1$ \\
25 & $81.1$ & $69.1$ & $75.7$ & $83.3$ & $90.7$ \\
50 & $65.6$ & $64.0$ & $75.1$ & $79.8$ & $86.9$ \\
\bottomrule
\end{tabular}
\end{table}

\begin{table}[!t]
\color{diffcolor}
\caption{ Comparison of MCTS with and without random rollouts (till depth of 20) in Pommerman FFA against three rule-based baseline opponents on 400 games.}
\label{t:mcts-random}
\centering
\begin{tabular}{l|cc|cc} 
\toprule
 & \multicolumn{2}{c|}{MCTS w/ random rollouts} & \multicolumn{2}{c}{MCTS w/o random rollouts} \\
 & UCB & TS & UCB & TS \\
\midrule
Wins   & $37.3\%$ & $40.8\%$ & $21.0\%$ & $17.8\%$ \\
Draws  & $32.4\%$ & $36.1\%$ & $17.7\%$ & $14.2\%$ \\
Losses & $30.3\%$ & $23.1\%$ & $61.3\%$ & $68.0\%$ \\
\bottomrule
\end{tabular}
\end{table}

\begin{table}[!t]
\caption{Hyperparameters of oracle planner and follower network in Pommerman. We report all the values considered during random search and the final chosen values are highlighted.}
\label{t:pommerman-hp}
\centering
\begin{tabular}{lll}
\toprule
& Parameter & Values \\
\midrule
\multirow{3}{*}{Oracle}
& Planning Depth & $[10, 15, \mathbf{20}]$ \\
& Num. Iterations & $[50, \mathbf{100}]$ \\
\midrule
\multirow{2}{*}{Follower}
& Batch Size & $[\mathbf{32}, 64, 128]$ \\
& Learning Rate & $\mathbf{0.001}$ \\
\bottomrule
\end{tabular}
\end{table}

\begin{table}[!t]
\caption{Hyperparameters of oracle planner and follower network in Clash Royale. We report all the values considered during random search and the final chosen values are highlighted.}
\label{t:cr-hp}
\centering
\begin{tabular}{lll}
\toprule
& Parameter & Values \\
\midrule
\multirow{3}{*}{Oracle}
& Planning Depth & $[10, 25, \mathbf{50}]$ \\
& Num. Iterations & $4\max(|A_1(s)|, |A_2(s)|)$ \\
& Num. Positions & $[32, \mathbf{64}]$ \\
\midrule
\multirow{4}{*}{Follower}
& Batch Size & $[\mathbf{32}, 64, 128]$ \\
& Learning Rate & $[0.001, \mathbf{0.0003}]$ \\
& Embedding Size & $[\mathbf{32}, 64]$ \\
& Hidden Size & $[64, \mathbf{128}, 256]$ \\
\bottomrule
\end{tabular}
\end{table}


%



\section*{Acknowledgment}

The authors would like to thank Steven Spencer, Hotloo Xiranood, Mika Seppä and everybody else at Supercell for fruitful discussions, comments on the draft of this paper, computational infrastructure, manual testing of learned agents and other support.

\ifCLASSOPTIONcaptionsoff
  \newpage
\fi



\bibliographystyle{IEEEtran}
\bibliography{IEEEabrv, bibliography}

\begin{thebibliography}{10}
\providecommand{\url}[1]{#1}
\csname url@samestyle\endcsname
\providecommand{\newblock}{\relax}
\providecommand{\bibinfo}[2]{#2}
\providecommand{\BIBentrySTDinterwordspacing}{\spaceskip=0pt\relax}
\providecommand{\BIBentryALTinterwordstretchfactor}{4}
\providecommand{\BIBentryALTinterwordspacing}{\spaceskip=\fontdimen2\font plus
\BIBentryALTinterwordstretchfactor\fontdimen3\font minus
  \fontdimen4\font\relax}
\providecommand{\BIBforeignlanguage}[2]{{%
\expandafter\ifx\csname l@#1\endcsname\relax
\typeout{** WARNING: IEEEtran.bst: No hyphenation pattern has been}%
\typeout{** loaded for the language `#1'. Using the pattern for}%
\typeout{** the default language instead.}%
\else
\language=\csname l@#1\endcsname
\fi
#2}}
\providecommand{\BIBdecl}{\relax}
\BIBdecl

\bibitem{silver2017mastering}
D.~Silver, J.~Schrittwieser, K.~Simonyan, I.~Antonoglou, A.~Huang, A.~Guez,
  T.~Hubert, L.~Baker, M.~Lai, A.~Bolton \emph{et~al.}, ``Mastering the game of
  {Go} without human knowledge,'' \emph{Nature}, vol. 550, no. 7676, pp.
  354--359, 2017.

\bibitem{moravvcik2017deepstack}
M.~Morav{\v{c}}{\'\i}k, M.~Schmid, N.~Burch, V.~Lis{\`y}, D.~Morrill, N.~Bard,
  T.~Davis, K.~Waugh, M.~Johanson, and M.~Bowling, ``Deepstack: Expert-level
  artificial intelligence in heads-up no-limit {Poker},'' \emph{Science}, vol.
  356, no. 6337, pp. 508--513, 2017.

\bibitem{brown2018superhuman}
N.~Brown and T.~Sandholm, ``Superhuman {AI} for heads-up no-limit {Poker}:
  Libratus beats top professionals,'' \emph{Science}, vol. 359, no. 6374, pp.
  418--424, 2018.

\bibitem{brown2019superhuman}
------, ``Superhuman {AI} for multiplayer {Poker},'' \emph{Science}, vol. 365,
  no. 6456, pp. 885--890, 2019.

\bibitem{vinyals2019grandmaster}
O.~Vinyals, I.~Babuschkin, W.~M. Czarnecki, M.~Mathieu, A.~Dudzik, J.~Chung,
  D.~H. Choi, R.~Powell, T.~Ewalds, P.~Georgiev \emph{et~al.}, ``Grandmaster
  level in {StarCraft II} using multi-agent reinforcement learning,''
  \emph{Nature}, vol. 575, no. 7782, pp. 350--354, 2019.

\bibitem{berner2019dota}
C.~Berner, G.~Brockman, B.~Chan, V.~Cheung \emph{et~al.}, ``Dota~2 with large
  scale deep reinforcement learning,'' \emph{arXiv preprint arXiv:1912.06680},
  2019.

\bibitem{DBLP:journals/corr/abs-1809-07124}
C.~Resnick, W.~Eldridge, D.~Ha, D.~Britz, J.~Foerster, J.~Togelius, K.~Cho, and
  J.~Bruna, ``Pommerman: {A} multi-agent playground,'' \emph{CoRR}, vol.
  abs/1809.07124, 2018.

\bibitem{brown2017safe}
N.~Brown and T.~Sandholm, ``Safe and nested subgame solving for
  imperfect-information games,'' in \emph{Advances in neural information
  processing systems}, 2017, pp. 689--699.

\bibitem{frank1998search}
I.~Frank and D.~Basin, ``Search in games with incomplete information: A case
  study using {Bridge} card play,'' \emph{Artificial Intelligence}, vol. 100,
  no. 1-2, pp. 87--123, 1998.

\bibitem{ginsberg2001gib}
M.~L. Ginsberg, ``Gib: Imperfect information in a computationally challenging
  game,'' \emph{Journal of Artificial Intelligence Research}, vol.~14, pp.
  303--358, 2001.

\bibitem{bjarnason2009lower}
R.~Bjarnason, A.~Fern, and P.~Tadepalli, ``Lower bounding klondike solitaire
  with monte-carlo planning,'' in \emph{Nineteenth International Conference on
  Automated Planning and Scheduling}, 2009.

\bibitem{cowling2012information}
P.~I. Cowling, E.~J. Powley, and D.~Whitehouse, ``Information set {Monte Carlo}
  tree search,'' \emph{IEEE Transactions on Computational Intelligence and AI
  in Games}, vol.~4, no.~2, pp. 120--143, 2012.

\bibitem{hansen2004dynamic}
E.~A. Hansen, D.~S. Bernstein, and S.~Zilberstein, ``Dynamic programming for
  partially observable stochastic games,'' 2004.

\bibitem{matiisen2018pommerman}
T.~Matiisen, ``Pommerman baselines,''
  \url{https://github.com/tambetm/pommerman-baselines}, 2018.

\bibitem{anthony2019policy}
T.~Anthony, R.~Nishihara, P.~Moritz, T.~Salimans, and J.~Schulman, ``Policy
  gradient search: Online planning and expert iteration without search trees,''
  \emph{NeurIPS 2018 Deep Reinforcement Learning Workshop}, 2019.

\bibitem{browne2012survey}
C.~B. Browne, E.~Powley, D.~Whitehouse, S.~M. Lucas, P.~I. Cowling,
  P.~Rohlfshagen, S.~Tavener, D.~Perez, S.~Samothrakis, and S.~Colton, ``A
  survey of {Monte-Carlo} tree search methods,'' \emph{IEEE Transactions on
  Computational Intelligence and AI in games}, vol.~4, no.~1, pp. 1--43, 2012.

\bibitem{auer2002finite}
P.~Auer, N.~Cesa-Bianchi, and P.~Fischer, ``Finite-time analysis of the
  multiarmed bandit problem,'' \emph{Machine learning}, vol.~47, no. 2-3, pp.
  235--256, 2002.

\bibitem{thompson1933likelihood}
W.~R. Thompson, ``On the likelihood that one unknown probability exceeds
  another in view of the evidence of two samples,'' \emph{Biometrika}, vol.~25,
  no. 3/4, pp. 285--294, 1933.

\bibitem{perick2012comparison}
P.~Perick, D.~L. St-Pierre, F.~Maes, and D.~Ernst, ``Comparison of different
  selection strategies in {Monte-Carlo} tree search for the game of tron,'' in
  \emph{2012 IEEE Conference on Computational Intelligence and Games
  (CIG)}.\hskip 1em plus 0.5em minus 0.4em\relax IEEE, 2012, pp. 242--249.

\bibitem{perez2019analysis}
D.~Perez-Liebana, R.~D. Gaina, O.~Drageset, E.~Ilhan, M.~Balla, and S.~M.
  Lucas, ``Analysis of statistical forward planning methods in {Pommerman},''
  in \emph{Proceedings of the AAAI Conference on Artificial Intelligence and
  Interactive Digital Entertainment}, vol.~15, no.~1, 2019, pp. 66--72.

\bibitem{perez2013rolling}
D.~Perez, S.~Samothrakis, S.~Lucas, and P.~Rohlfshagen, ``Rolling horizon
  evolution versus tree search for navigation in single-player real-time
  games,'' in \emph{Proceedings of the 15th annual conference on Genetic and
  evolutionary computation}, 2013, pp. 351--358.

\bibitem{ross2011reduction}
S.~Ross, G.~Gordon, and D.~Bagnell, ``A reduction of imitation learning and
  structured prediction to no-regret online learning,'' in \emph{Proceedings of
  the fourteenth international conference on artificial intelligence and
  statistics}, 2011, pp. 627--635.

\bibitem{gao2019skynet}
C.~Gao, P.~Hernandez-Leal, B.~Kartal, and M.~E. Taylor, ``Skynet: A top deep
  {RL} agent in the inaugural {Pommerman} team competition,'' \emph{arXiv
  preprint arXiv:1905.01360}, 2019.

\bibitem{adkb2018td}
A.~Amiranashvili, A.~Dosovitskiy, V.~Koltun, and T.~Brox, ``{TD} or not {TD}:
  Analyzing the role of temporal differencing in deep reinforcement learning,''
  in \emph{International Conference on Learning Representations (ICLR)}, 2018.

\bibitem{mnih2015human}
V.~Mnih, K.~Kavukcuoglu, D.~Silver, A.~A. Rusu, J.~Veness, M.~G. Bellemare,
  A.~Graves, M.~Riedmiller, A.~K. Fidjeland, G.~Ostrovski \emph{et~al.},
  ``Human-level control through deep reinforcement learning,'' \emph{Nature},
  vol. 518, no. 7540, p. 529, 2015.

\bibitem{ye2019mastering}
D.~Ye, Z.~Liu, M.~Sun, B.~Shi, P.~Zhao, H.~Wu, H.~Yu, S.~Yang, X.~Wu, Q.~Guo
  \emph{et~al.}, ``Mastering complex control in {MOBA} games with deep
  reinforcement learning,'' \emph{arXiv preprint arXiv:1912.09729}, 2019.

\bibitem{schulman2017proximal}
J.~Schulman, F.~Wolski, P.~Dhariwal, A.~Radford, and O.~Klimov, ``Proximal
  policy optimization algorithms,'' \emph{arXiv preprint arXiv:1707.06347},
  2017.

\bibitem{espeholt2018impala}
L.~Espeholt, H.~Soyer, R.~Munos, K.~Simonyan, V.~Mnih, T.~Ward, Y.~Doron,
  V.~Firoiu, T.~Harley, I.~Dunning \emph{et~al.}, ``{IMPALA}: Scalable
  distributed deep-{RL} with importance weighted actor-learner architectures,''
  in \emph{International Conference on Machine Learning}, 2018, pp. 1406--1415.

\bibitem{shafieicomparing}
M.~Shafiei, N.~Sturtevant, and J.~Schaeffer, ``Comparing {UCT} versus {CFR} in
  simultaneous games,'' \emph{IJCAI Workshop on General Game Playing}, 2009.

\bibitem{teytaud2011lemmas}
F.~Teytaud and O.~Teytaud, ``Lemmas on partial observation, with application to
  phantom games,'' in \emph{2011 IEEE Conference on Computational Intelligence
  and Games (CIG'11)}.\hskip 1em plus 0.5em minus 0.4em\relax IEEE, pp.
  243--249.

\bibitem{bovsansky2016algorithms}
B.~Bo{\v{s}}ansk{\`y}, V.~Lis{\`y}, M.~Lanctot, J.~{\v{C}}erm{\'a}k, and M.~H.
  Winands, ``Algorithms for computing strategies in two-player simultaneous
  move games,'' \emph{Artificial Intelligence}, vol. 237, pp. 1--40, 2016.

\bibitem{long2010understanding}
J.~R. Long, N.~R. Sturtevant, M.~Buro, and T.~Furtak, ``Understanding the
  success of perfect information {Monte Carlo} sampling in game tree search,''
  in \emph{Twenty-Fourth AAAI Conference on Artificial Intelligence}, 2010.

\bibitem{schafer2008uct}
J.~Sch{\"a}fer, M.~Buro, and K.~Hartmann, ``The {UCT} algorithm applied to
  games with imperfect information,'' 2008.

\bibitem{buro2009improving}
M.~Buro, J.~R. Long, T.~Furtak, and N.~Sturtevant, ``Improving state
  evaluation, inference, and search in trick-based card games,'' in
  \emph{Twenty-First International Joint Conference on Artificial
  Intelligence}, 2009.

\bibitem{liu2019}
T.~Liu, Z.~Zheng, H.~Li, K.~Bian, and L.~Song, ``Playing card-based {RTS} games
  with deep reinforcement learning,'' in \emph{Proceedings of the Twenty-Eighth
  International Joint Conference on Artificial Intelligence, {IJCAI-19}}.\hskip
  1em plus 0.5em minus 0.4em\relax International Joint Conferences on
  Artificial Intelligence Organization, 2019, pp. 4540--4546.

\bibitem{guo2014deep}
X.~Guo, S.~Singh, H.~Lee, R.~L. Lewis, and X.~Wang, ``Deep learning for
  real-time {Atari} game play using offline {Monte-Carlo} tree search
  planning,'' in \emph{Advances in neural information processing systems},
  2014, pp. 3338--3346.

\bibitem{gai2010learning}
Y.~Gai, B.~Krishnamachari, and R.~Jain, ``Learning multiuser channel
  allocations in cognitive radio networks: A combinatorial multi-armed bandit
  formulation,'' in \emph{2010 IEEE Symposium on New Frontiers in Dynamic
  Spectrum (DySPAN)}.\hskip 1em plus 0.5em minus 0.4em\relax IEEE, 2010, pp.
  1--9.

\bibitem{chen2013combinatorial}
W.~Chen, Y.~Wang, and Y.~Yuan, ``Combinatorial multi-armed bandit: General
  framework and applications,'' in \emph{International Conference on Machine
  Learning}, 2013, pp. 151--159.

\bibitem{ontanon2013combinatorial}
S.~Ontan{\'o}n, ``The combinatorial multi-armed bandit problem and its
  application to real-time strategy games,'' in \emph{Proceedings of the Ninth
  AAAI Conference on Artificial Intelligence and Interactive Digital
  Entertainment}, 2013, pp. 58--64.

\bibitem{ontanon2017combinatorial}
------, ``Combinatorial multi-armed bandits for real-time strategy games,''
  \emph{Journal of Artificial Intelligence Research}, vol.~58, pp. 665--702,
  2017.

\bibitem{zinkevich2008regret}
M.~Zinkevich, M.~Johanson, M.~Bowling, and C.~Piccione, ``Regret minimization
  in games with incomplete information,'' in \emph{Advances in neural
  information processing systems}, 2008, pp. 1729--1736.

\bibitem{blackburn2014cheating}
J.~Blackburn, N.~Kourtellis, J.~Skvoretz, M.~Ripeanu, and A.~Iamnitchi,
  ``Cheating in online games: A social network perspective,'' \emph{ACM
  Transactions on Internet Technology (TOIT)}, vol.~13, no.~3, pp. 1--25, 2014.

\bibitem{zuo2016bad}
X.~Zuo, C.~Gandy, J.~Skvoretz, and A.~Iamnitchi, ``Bad apples spoil the fun:
  Quantifying cheating in online gaming,'' in \emph{Tenth International AAAI
  Conference on Web and Social Media}, 2016.

\bibitem{paay2018motivations}
J.~Paay, J.~Kjeldskov, D.~Internicola, and M.~Thomasen, ``Motivations and
  practices for cheating in pok{\'e}mon go,'' in \emph{Proceedings of the 20th
  International Conference on Human-Computer Interaction with Mobile Devices
  and Services}, 2018, pp. 1--13.

\end{thebibliography}

%








\end{document}